\documentclass{article}
\usepackage{graphicx} 

\usepackage{etoc}
\usepackage[preprint]{neurips_2024}
\usepackage{graphicx}
\usepackage{booktabs}
\usepackage[table,xcdraw]{xcolor}
\usepackage{wrapfig}
\usepackage{caption} 
\usepackage{tocloft}
\usepackage{natbib}
\usepackage[utf8]{inputenc} 
\usepackage[T1]{fontenc}    
\usepackage{hyperref}       
\usepackage{url}            
\usepackage{booktabs}       
\usepackage{amsfonts}       
\usepackage{amsmath}
\usepackage{mathtools}
\usepackage{nicefrac}       
\usepackage{microtype}      
\usepackage{xcolor}         
\usepackage{amsthm}
\usepackage{multirow}
\usepackage{algorithm}
\usepackage{algorithmic}
\usepackage{wrapfig}
\usepackage{subcaption}
\usepackage{caption} 
\usepackage{tocloft}
\usepackage{amssymb}
\usepackage{fontawesome5}
\usepackage{mathrsfs}

\newcounter{ToDo}
\newcounter{gaocomm} 
\newcounter{Note}
\definecolor{blue-violet}{rgb}{0.00,0.75,0.90}
\definecolor{mygreen}{rgb}{0.0, 0.5, 0.0}
\definecolor{awesome}{rgb}{1.0, 0.13, 0.32}
\definecolor{bostonuniversityred}{rgb}{1.0, 0.0, 0.0}

\newcommand{\cmark}{\ding{51}} 
\newcommand{\xmark}{\ding{55}} 
\newcommand{\semitick}{{\cmark}\kern-0.62em{\xmark}}

\definecolor{customblue}{HTML}{b9dcb6}  

\hypersetup{colorlinks=true,linkcolor=red!70!black,linktocpage=false,citebordercolor=blue!70!black,citecolor=blue!70!black,anchorcolor=blue!70!black}

\title{Wiener Chaos Expansion based Neural Operator for Singular Stochastic Partial Differential Equations}

\author{
  Dai Shi  \thanks{\faEnvelope \ Corresponding to \texttt{ds2213@cam.ac.uk}.}\\
  University of Cambridge\\
  \And
  Luke Thompson \\
  University of Sydney\\
  \And 
  Andi Han \\ 
  University of Sydney \\
  \AND
  Peiyan Hu \\
  Chinese Academy of Science \\
 \And
  Junbin Gao \\
  University of Sydney \\
  \And
  José Miguel Hernández-Lobato\\
  University of Cambridge\\
}

\begin{document}

\maketitle

\begin{center}
\textit{For one of the award-winning models from the SPDE modeling Competition}\footnotemark
\end{center}
\footnotetext{\url{https://hackathon2.deepintomlf.ai/competitions/92/}}

\begin{abstract}
In this paper, we explore how our recently developed Wiener Chaos Expansion (WCE)-based neural operator (NO) can be applied to singular stochastic partial differential equations, e.g., the dynamic $\boldsymbol{\Phi}^4_2$ model simulated in the recent works. 
Unlike the previous WCE-NO 
which solves SPDEs by simply inserting Wick-Hermite features into the backbone NO model, we leverage feature-wise linear modulation (FiLM) to appropriately capture the dependency between the solution of singular SPDE and its smooth remainder.  The resulting WCE-FiLM-NO shows excellent performance on $\boldsymbol{\Phi}^4_2$, as measured by relative $L_2$ loss, out-of-distribution $L_2$ loss, and autocorrelation score; all without the help of renormalisation factor. In addition, we also show the potential of simulating $\boldsymbol{\Phi}^4_3$ data, which is more aligned with real scientific practice in statistical quantum field theory. To the best of our knowledge, this is among the first works to develop an efficient data-driven surrogate for the dynamical $\boldsymbol{\Phi}^4_3$ model \footnote{{Our code is available upon request.}}.
\end{abstract}

\section{Introduction}
Stochastic partial differential equations serve as one of the fundamental tools for modelling spatially distributed dynamical systems under uncertainty across a wide range of domains \citep{hairer2009introduction}, including atmospheric and climate science \citep{hasselmann1976stochastic,palmer2019stochastic}, physics (e.g., turbulent flows and quantum field theory \citep{duch2025flow,hairer2015regularity}), biology \citep{sabbar2026selective,wilkinson2018stochastic}, and economics/finance \citep{barone1987efficient}. Neural operators (NO) \citep{kovachki2023neural}, as remarkable tools for approximating the mapping between functions, have been applied to learn the solutions of various differential equations, e.g., ODE \citep{lu2019deeponet}, PDE  \citep{li2020fourier,hu2024wavelet}, SDE \citep{eigel2024functional}, and SPDE \citep{salvi2022neural,shi2026expanding,neufeld2024solving}. While models such as NSPDE \citep{salvi2022neural}, NORS \citep{hu2022neural}, DLR-NET \citep{gong2023deep}, and SDENO \citep{shi2026expanding} have shown excellent numerical results in different SPDEs, their performances on singular SPDEs remain unknown.  One of the major reasons for this limitation is the lack of efficient and reliable simulation pipelines for singular SPDEs, as their well-posedness and numerical stability typically rely on delicate renormalisation procedures and counterterms, together with fine discretisation and careful resolution control \citep{zhu2018lattice,berglund2022introduction}. Another reason might be due to the inefficiency of the traditional solvers, as they usually solve SPDEs in an iterative way, and non-convergence of the classic NO baselines (e.g., FNO \citep{li2020fourier}) in singular SPDEs. Thanks to the recent work in \citep{li2025spdebench} where an efficient non-singular SPDE, e.g., dynamic $\boldsymbol{\Phi}^4_2$ dataset is developed based on the renormalization method, and several tests are conducted in both renormalized and non-renormalized dataset versions. 

In this paper, we present a prize-winning model that was selected as one of the winners in the singular SPDE modelling competition. Our model is developed based on the classic Wiener Chaos Expansion (WCE) theory \citep{luo2006wiener,lototsky2006wiener,lototsky2017stochastic}, which states that the solution of semi-linear SPDEs can be expressed by a linear combination of propagations that are governed by a group of deterministic PDEs and the related Wick-Hermite features (\eqref{eq:spde_wce} below). By supplying these Wick features with a specific order to the classic FNO, one can obtain the solution of the so-called \eqref{eq:shift_equation}, and we further conduct a parameterized affine transformation to the FNO output so that the overall solution of $\boldsymbol{\Phi}^4_2$ can be constructed based on Da Prato-Debussche decomposition (DPDD). Numerical results in the renormalized dataset show that our model not only outperforms NSPDE in different truncations of noise but also shows better generalizability in terms of cross-truncation evaluation (See Table~\ref{tab:phi42_results}) even without leveraging the renormalization factor. Finally, we also demonstrate a simulation pipeline for an even more singular SPDE, namely the dynamical $\boldsymbol{\Phi}^4_3$ model, which is closely related to stochastic quantisation in statistical field theory \citep{hairer2015regularity}. Although serving as an initial phototype, we hope our work could pave the path of utilizing deep learning tools to solve the complex singular SPDEs in the future. 

\section{Background}

\subsection{WCE for SPDEs}


Let $T>0$ and let $(\Omega,\mathcal F,(\mathcal F_t)_{t\in[0,T]},\mathbb P)$ be a filtered probability space. Let $\mathcal H$ and $\widetilde{\mathcal H}$ be separable Hilbert spaces. We consider the semi-linear SPDE
\begin{align}\label{eq:spde_initial}
dX_t = \big(AX_t + F(t,X_t)\big),dt + B(t,X_t),dW_t,\qquad X_0=\chi_0\in\mathcal H, \tag{SPDE}
\end{align}
where $A:\mathcal H\to\mathcal H$ is a (possibly unbounded) linear operator, $F:[0,T]\times\Omega\times\mathcal H\to\mathcal H$, and $B:[0,T]\times\Omega\times\mathcal H\to L_2(\widetilde{\mathcal H};\mathcal H)$ are measurable maps, and $\chi_0$ is deterministic. The driving noise $W=(W_t)_{t\in[0,T]}$ is a Q-Brownian motion on $\widetilde{\mathcal H}$ adapted to $(\mathcal F_t)$.
Based on the WCE theory \citep{kalpinelli2011wiener,lototsky2006stochastic,lototsky2006wiener,mikulevicius1998linear}, when \eqref{eq:spde_initial} has unique mild solution under Lipschitz \& linear-growth assumptions \citep{da2014stochastic}, its solution can be written as 
\begin{align}\label{eq:spde_wce}
    X(t,\omega) = \sum_{\alpha \in \mathcal{J}} z_\alpha(t) \xi_\alpha(\omega),  \tag{WCE-SPDE}
\end{align}

for $\omega \in \Omega$, and 
$\{\xi_\alpha\}_{\alpha\in\mathcal J}$, with $\mathcal{J}=\{\alpha=(\alpha_{ij})_{i,j\in\mathbb{N}}\in\mathbb{N}_0^{\mathbb{N}\times \mathbb{N}}: |\alpha|:=\sum_{i,j}\alpha_{ij}<\infty\}$,
is the Wick-Hermite basis (or Wick features) induced by $W$. Specifically, let $\{e_j\}_{j\in\mathbb N}$ be an orthonormal basis of $L^2([0,T])$, one can obtain Gaussian random variables as
$\xi_{ij}:=\int_0^T e_j(s)\,dW^{(i)}_s$. Then for any multi-index $\alpha=(\alpha_{ij})$ with finite support, the Wick features $\xi_\alpha$ can be computed by
$\xi_\alpha:=\frac{1}{\sqrt{\alpha!}}\prod_{i,j} h_{\alpha_{ij}}(\xi_{ij})$, where $h_k$ is the $k$-th Hermite polynomial, i.e., $h_k(x) = (-1)^k e^{x^2/2} \frac{d^k}{dx^k} e^{-x^2/2}$, i.e., $h_0(x) = 1, \,\, h_1(x) = x, \,\, h_2(x) = x^2 - 1, \,\, h_3(x) = x^3 - 3x$, and $\alpha!\coloneqq\prod_{m,j}(\alpha_{mj}!)$. Furthermore, let $I, J, K$ limit  the number of noise components $\{W^{(i)}\}_{i=1}^I$, the number of temporal modes $\{e_j\}_{j=1}^J$ and the maximal order of the Hermite polynomials (introduce latter), respectively,  for $I,J,K\in\mathbb N$, one can define the truncated index set
\begin{align}\label{eq:J_IJK}
\mathcal J_{I,J,K}
\;:=\;
\Bigl\{
\alpha=(\alpha_{ij})_{i,j\in\mathbb N}\in \mathbb N_0^{\mathbb N\times\mathbb N}
:\ \alpha_{ij}=0\ \text{if } i>I\ \text{or } j>J,\ \ |\alpha|\le K
\Bigr\}
\ \subset\ \mathcal J.
\end{align}
We note that the condition $|\alpha|\coloneqq \sum_{i,j}\alpha_{ij} \leq K$ is \textbf{crucial} as it retains chaos indices that grow combinatorially, e.g., $|\mathcal J_{I,J,K}|
\;=\;
\sum_{k=0}^{K} \binom{IJ+k-1}{k}
=
\binom{IJ+K}{K}
=
\frac{(IJ+K)!}{(IJ)!\,K!}$, as illustrated in \citep{neufeld2024solving,luo2006wiener}. As in this work, we only consider one-dimensional Q-Brownian motion in two/three-dimensional space (e.g., $\boldsymbol{\Phi}^4_2$ and $\boldsymbol{\Phi}^4_3$), we will simply set $I =1$, thus the combination reduces to $\frac{(J+K)!}{J!\, K!}$.

More importantly, based on the WCE theory \citep{neufeld2024solving,luo2006wiener}, the coefficients, i.e., $\{z_\alpha(t)\}_{\alpha\in \mathcal J}$, also known as the propagators \citep{huschto2014solving}, 
are determined by a group of \textit{deterministic PDEs}. This serves as the main motivation for the recent WCE-based SDE and SPDE NO models \citep{eigel2024functional,shi2026expanding,zeng2026latent}, where the elements in $\{z_\alpha(t)\}_{\alpha\in \mathcal J}$ are parameterised by the learnable neural networks.

\subsection{WCE based NO}
NOs based on the WCE theory have been applied to both SDE and SPDE realms. In the SDE case, \cite{eigel2024functional} utilizes deep neural networks (e.g., MLPs) for approximating the propagators that are determined by ODEs. For the SPDE case, the initial numerical method was developed by \cite{neufeld2024solving}, in which $z(t)$ is obtained by either deep or random neural networks. Motivated by these works, \cite{shi2026expanding} proposed an unified WCE-NO framework for both SDE and SPDE, and applied their model into various tasks, such as diffusion image generation, topological interpolation, financial time series extrapolation, $\boldsymbol{\Phi}^4_1$ and Naiver-Stokes equation. However, it is still unclear whether WCE based NO model can approximate the solution of high-dimensional \textit{singular} SPDEs. 

\subsection{Dynamic $\boldsymbol{\Phi}^4_2$, Renormalization, and Initial Numerical Solution}
As a specific case of \eqref{eq:spde_initial}, the dynamic $\boldsymbol{\Phi}^4_2$ can be written as 
\begin{align}\label{eq:phi42_spde_W}
\begin{cases}
du(t) = \bigl(\Delta u(t) - u(t)^3\bigr)\,dt + \sigma\, dW_t, & t\in[0,T],\\
u(0)=u_0,
\end{cases} \tag{$\boldsymbol{\Phi}^4_2$}
\end{align}
where $u(t)=u(t,\cdot)\in \mathcal H$ with $\mathcal H=L^2(\mathbb T^2)$, and $\mathbb{T} = \mathbb{R}/\mathbb{Z}$.
We refer the readers to \citep{gubinelli2019global} for a detailed mathematical treatment of the $\boldsymbol{\Phi}^4_d$ family, while the full simulation and renormalization details used in our experiments can be found in \citep{li2025spdebench}.

\paragraph{Renormalization}

Since the cubic nonlinearity is ill-defined under space-time white noise, the equation is rigorously constructed as the limit of regularized approximations. To this end, we first define the stochastic convolution $X_\epsilon$ as the solution to the linear heat equation driven by the truncated noise $W_\epsilon$:
\begin{align}\label{eq:X_epsilon_def}
dX_\epsilon(t) = \Delta X_\epsilon(t) dt + \sigma dW_\epsilon(t), \quad X_\epsilon(0) = 0.
\end{align}
Here, $W_\epsilon := \mathcal{P}_\epsilon W$ represents the spectral Galerkin approximation \citep{gubinelli2019global}, where $\mathcal{P}_\epsilon$ is a smoothing/projection operator that removes spatial Fourier frequencies above the cutoff scale $\epsilon$.
The renormalization constant is defined as the pointwise variance $a_\epsilon(t) := \mathbb{E}[|X_\epsilon(t)|^2]$. To regularize the singular nonlinearity, the classical cubic term is replaced by the Wick power $u^{\diamond 3}_\epsilon$, defined explicitly by subtracting the divergent counter-term:
\begin{align}\label{eq:wick_def}
u^{\diamond 3}_\epsilon(t) := u^3_\epsilon(t) - 3 a_\epsilon(t) u_\epsilon(t),
\end{align}
Consequently, the regularized approximation $u_\epsilon$ satisfies:
\begin{align}\label{eq:phi42_approx}
du_\epsilon(t) = \bigl(\Delta u_\epsilon(t) - u^{\diamond 3}_\epsilon(t)\bigr)\,dt + \sigma\, dW_\epsilon(t), \quad u_\epsilon(0) = \mathcal{P}_\epsilon u_0.
\end{align}

\paragraph{Initial Numerical Solution}
Directly solving Eq.~\eqref{eq:phi42_approx} is numerically unstable because $a_\epsilon \to \infty$ as $\epsilon \to 0$. To provide numerical solutions for the supervision of the potential models, one can choose to adopt the Da Prato-Debussche decomposition (DPDD) by setting $u_\epsilon = v_\epsilon + X_\epsilon$ \citep{cao2023yang,oh2021comparing}. 
Substituting the DPDD into Eq.~\eqref{eq:phi42_approx}, and expanding the renormalized cubic term using Eq.~\eqref{eq:wick_def}, one obtains the so-called \textit{Shift Equation} for $v_\epsilon$ (see \citep{li2025spdebench} for a detailed derivation):
\begin{align}\label{eq:shift_equation}
dv_\epsilon(t)
=\Delta v_\epsilon(t)\,dt
-
\bigl( v_\epsilon^3(t) + 3v_\epsilon^2(t) X_\epsilon(t) + 3v_\epsilon(t) X_\epsilon^{\diamond 2}(t) + X_\epsilon^{\diamond 3}(t) \bigr)\,dt
 ,
\tag{Shift Equation}
\end{align}
where the explicit infinite variance $a_\epsilon$ is algebraically \textit{absorbed} into the Wick powers of $X_\epsilon$, e.g., $X_\epsilon^{\diamond 2} := X_\epsilon^2 - a_\epsilon$ and $X_\epsilon^{\diamond 3} := X_\epsilon^3 - 3a_\epsilon X_\epsilon$. One can see that DPDD transfers an unstable equation in \eqref{eq:phi42_approx} into the combination of $X_\epsilon$ and a smooth remainder $v_\epsilon$ that solves the \eqref{eq:shift_equation}, which serves as the targets for the potential models. 

\section{Model Motivation and Architecture}
\begin{figure}[t]
    \centering
\includegraphics[width=1\linewidth]{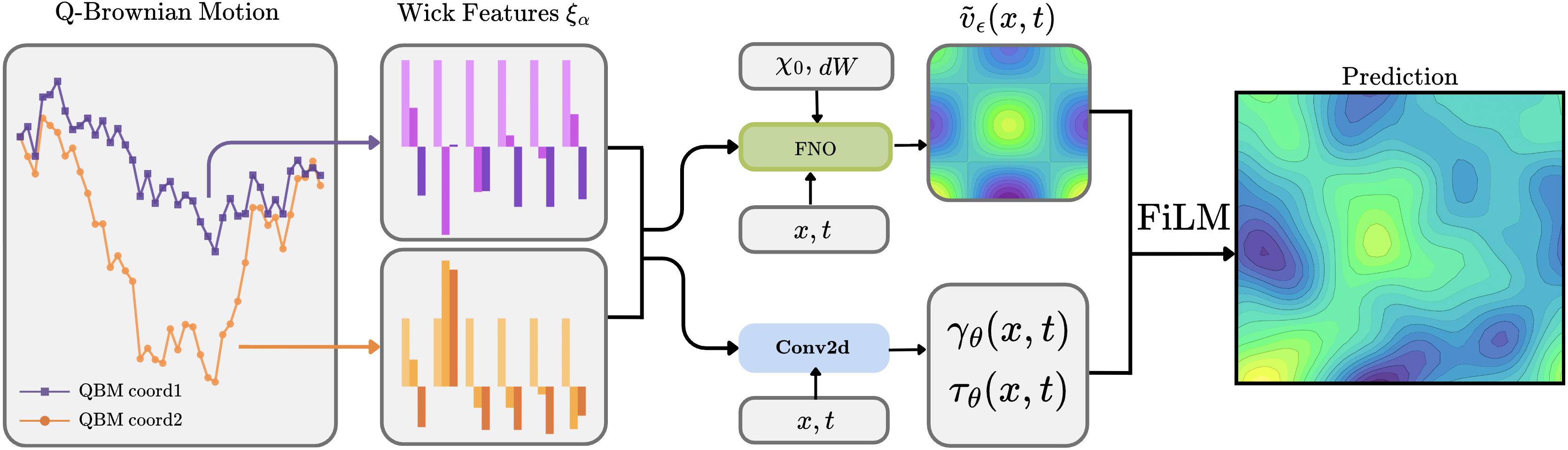}
    \caption{Architecture of WCE-FiLM-NO. Initial Wick features are computed through the Brownian increments, and then serve as the input to both FNO and FiLM computation to compute the remainder $\widehat{v}_\epsilon(t,x)$ and scaling/shifting coefficients (i.e., $\gamma, \tau$). Finally, the prediction is obtained by conducting the affine transformation between the output of FNO and Conv2D.}
    \label{fig:model_architecture}
\end{figure}

Based on the renormalization procedure above, to solve $v_\epsilon$, which is often known as the smooth remainder,  one needs access to the stochastic driving terms $X_\epsilon$, $X_\epsilon^{\diamond 2}$, and $X_\epsilon^{\diamond 3}$, after which the full field can be reconstructed via $u_\epsilon = v_\epsilon + X_\epsilon$. Fortunately, these driving terms are fully representable in terms of the first three chaos levels induced by the noise. That is 
$
X_\epsilon(t,x) = \sum_{|\alpha|=1} c_{\alpha,\epsilon}^{(1)}(t,x)\,\xi_\alpha, \,\,
X_\epsilon^{\diamond 2}(t,x) = \sum_{|\alpha|=2} c_{\alpha,\epsilon}^{(2)}(t,x)\,\xi_\alpha, \,\,\, \text{and} \,\,\, 
X_\epsilon^{\diamond 3}(t,x) = \sum_{|\alpha|=3} c_{\alpha,\epsilon}^{(3)}(t,x)\,\xi_\alpha, 
$
as one can verify $\widetilde{\mathcal H}_k  = \mathrm{span}\{\xi_\alpha, |\alpha| = k\}$.
This observation directly motivates our architecture: we feed the model with Wick features $\{\xi_\alpha\}_{\alpha \in \mathcal J}$ up to order $K=3$. The network can therefore use these Wick features to learn the smooth remainder $v_\epsilon$. This can be achieved by supplying $\xi_\alpha$ to those classic NO models such as FNO \citep{li2020fourier}, where coefficients, e.g., $c_{\alpha, \epsilon}(t,x)$ are learned implicitly in the Fourier space. Let $\widehat{v}_\epsilon(x,t)$ be the output of FNO, followed by DPDD, we further apply an affine transformation on $\widehat{v}_\epsilon(x,t)$, with the purpose of adding $X_\epsilon$ to $\widehat{v}_\epsilon(t,x)$,
that is 
\begin{align}\label{eq:model_film}
     \widehat{u}_\epsilon(t,x) = \bigl(1+\gamma_\theta(t,x)\bigr)\odot \widehat{v}_\epsilon(t,x) + \tau_\theta(t,x),
\end{align}
where $\widehat{u}_\epsilon(t,x)$ denotes the final prediction.
The modulation fields $(\gamma_\theta,\tau_\theta)$ are produced by a learnable conditioning network $g_\theta$
that processes Wick features together with spatial coordinates $x$ and time $t$,
\begin{align}
(\gamma_\theta,\tau_\theta) = g_\theta\!\left(\mathrm{Concat}\big[\mathcal \xi_\alpha,x,t\big]\right),
\end{align}
where $g_\theta$ is implemented by a simple $\mathrm{Conv2d}$. 
Figure~\ref{fig:model_architecture} shows the general process of our proposed model. Given this affine transformation spirit aligns with the feature-wise linear modulation proposed in \citep{perez2018film}, we name our model as WCE-FiLM-NO.

\section{Numerical Experiment}
We test our WCE-FiLM-NO on $\boldsymbol{\Phi}^4_2$ simulated from the work in \citep{li2025spdebench}. The relative $L_2$ errors are presented in Table~\ref{tab:phi42_results}. We compare our model with NSPDE developed in \citep{salvi2022neural} under four schemes, 
\begin{itemize}
    \item $W \mapsto u$ which assumes $W$ is observed but the initial condition is fixed all the time;

    \item $(W,a_\epsilon) \mapsto u$ in which $a_\epsilon$ is supplied to the model; 

    \item $(W, u_0) \mapsto u$ where $W$ is observed and $u_0$ changed across the samples; 

    \item $(W,u_0, a_\epsilon)$ where model leverages the full information.
\end{itemize}
We highlight that we don't need $a_\epsilon$ in WCE-FiLM-NO. The model is trained via two different truncations of noise, e.g., $\epsilon = 2$ and $\epsilon = 128$,
and cross evaluation is conducted after the model is trained. That is, the model is trained with $\epsilon =2$ but evaluated with $\epsilon = 128$. We note that we did not evaluate our model under the truncation number in between as presented in \citep{li2025spdebench}. In addition, we also did not present the results for other baselines that are designed for singular SPDEs, such as NORS \citep{hu2022neural} and DLR-Net \citep{gong2023deep}; we leave these for future work. 

\begin{table}[t]
\centering
\small
\setlength{\tabcolsep}{6pt}
\renewcommand{\arraystretch}{1.15}
\caption{Results in relative $L_2$ error in $\boldsymbol{\Phi}^4_2$.} \label{tab:phi42_results}
\begin{tabular}{l cc cc cc cc}
\toprule
\multirow{2}{*}{\textbf{Train set/Test set}} &
\multicolumn{2}{c}{\(W \mapsto u\)} &
\multicolumn{2}{c}{\((W,a_\varepsilon) \mapsto u\)} &
\multicolumn{2}{c}{\((W,u_0) \mapsto u\)} &
\multicolumn{2}{c}{\((W,u_0,a_\varepsilon) \mapsto u\)}\\
\cmidrule(lr){2-3}\cmidrule(lr){4-5}\cmidrule(lr){6-7}\cmidrule(lr){8-9}
& \(\mathcal{D}^{\mathrm{re}}_{J}\) & \(\mathcal{D}^{\mathrm{re}}_{128}\) &
  \(\mathcal{D}^{\mathrm{re}}_{J}\) & \(\mathcal{D}^{\mathrm{re}}_{128}\) &
  \(\mathcal{D}^{\mathrm{re}}_{J}\) & \(\mathcal{D}^{\mathrm{re}}_{128}\) &
  \(\mathcal{D}^{\mathrm{re}}_{J}\) & \(\mathcal{D}^{\mathrm{re}}_{128}\) \\
\midrule
\(\mathcal{D}^{\mathrm{re}}_{2} (\mathrm{NSPDE)}\)   & 0.010 & 4.087 & 0.034 & 0.998 & 0.010 & 2.805 & 0.042 & 0.998 \\
\(\mathcal{D}^{\mathrm{re}}_{128}\) & 0.230 & 0.230 & 0.028 & 0.028 & 0.223 & 0.223 & 0.029 & 0.029 \\
\hline
\(\mathcal{D}^{\mathrm{re}}_{2}(\text{WCE-FiLM-NO})\)   & \textbf{0.004} & \textbf{0.948} & -- & -- & \textbf{0.009} & \textbf{0.915} & -- & -- \\
\(\mathcal{D}^{\mathrm{re}}_{128}\) & \textbf{0.017} & \textbf{0.017} & -- & -- & $\textbf{0.025}$ & $\textbf{0.025}$ & -- & -- \\
\bottomrule
\end{tabular}
\end{table}

\paragraph{Results}

From Table~\ref{tab:phi42_results}, one can check that WCE-FiLM-NO outperforms NSPDE in all settings even without utilising the renormalisation factor, i.e., $a_\epsilon$. In terms of the cross-evaluation, although NSPDE can produce good results, these good results are with the help of $a_\epsilon$, e.g., 0.998 in the cases of $(W, a_\epsilon) \mapsto u$ and $(W,u_0,a_\epsilon) \mapsto u$. However, our model does not require the computation of $a_\epsilon$ and shows remarkable performance in the challenging scenarios, such as 0.948 in $W\mapsto u$ and $0.915$ in $(W, u_0) \mapsto u$, suggesting better generalizability. 

\section{Future work: $\boldsymbol{\Phi}^4_3$}
\begin{figure}[t]
    \centering
    \begin{subfigure}[t]{0.32\textwidth}
        \centering
        \includegraphics[width=\linewidth]{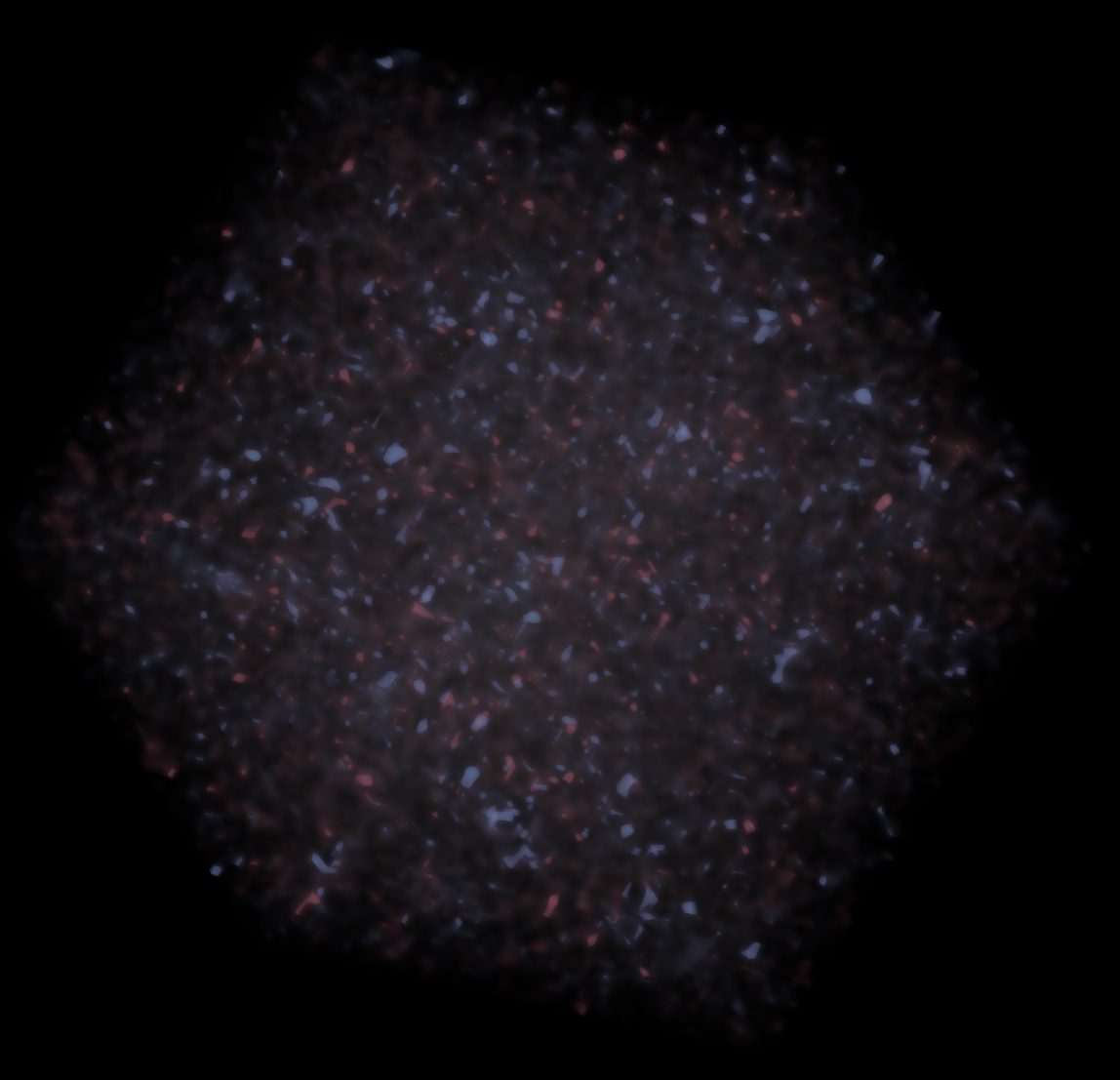}
        \caption{$\boldsymbol \Phi_3^4$ at $t=0$}
        \label{fig:phi43-a}
    \end{subfigure}
    \hfill
    \begin{subfigure}[t]{0.32\textwidth}
        \centering
        \includegraphics[width=\linewidth]{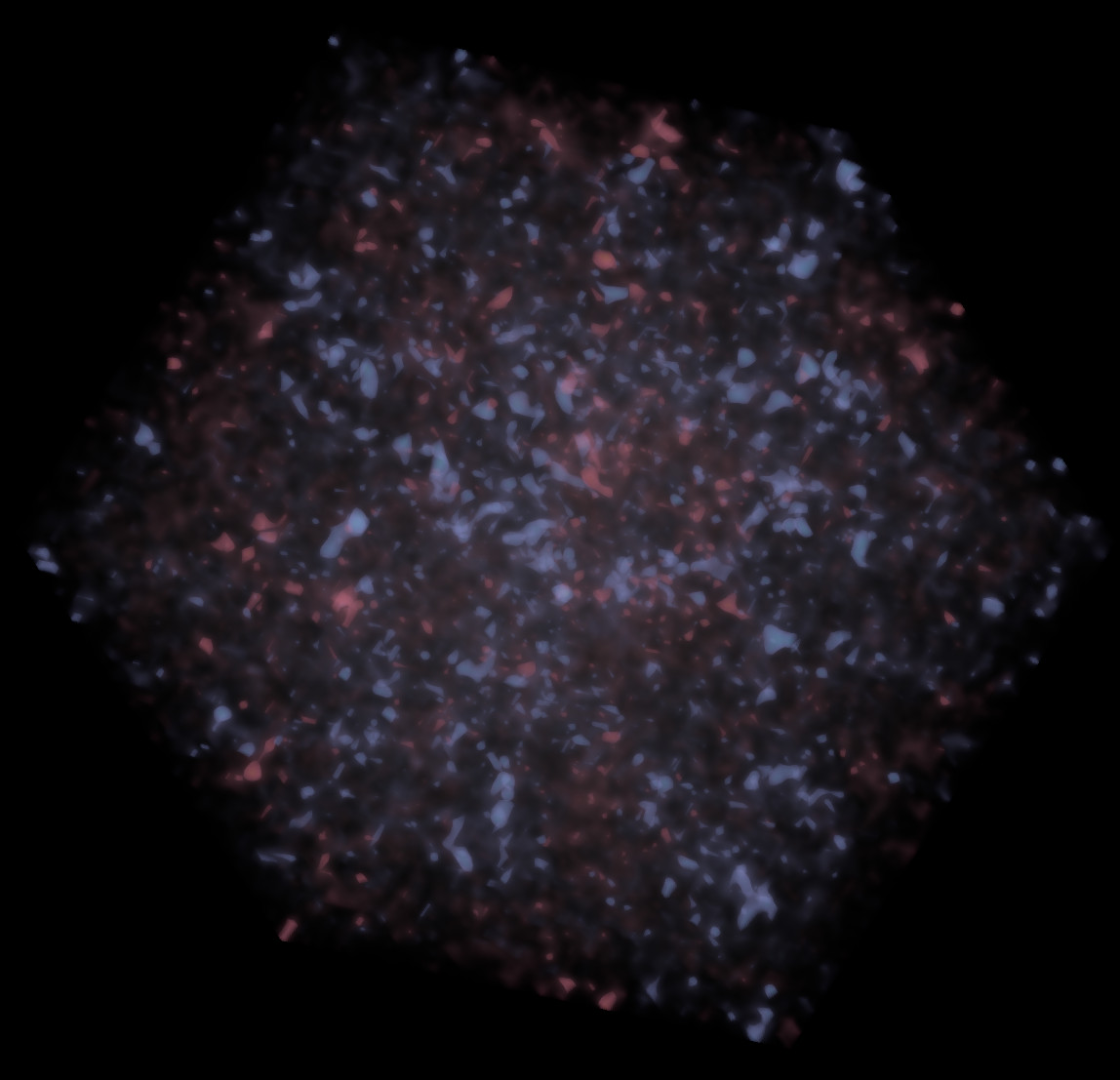}
        \caption{$ \boldsymbol \Phi_3^4$ at $t=0.5$}
        \label{fig:phi43-b}
    \end{subfigure}
    \hfill
    \begin{subfigure}[t]{0.32\textwidth}
        \centering
        \includegraphics[width=\linewidth]{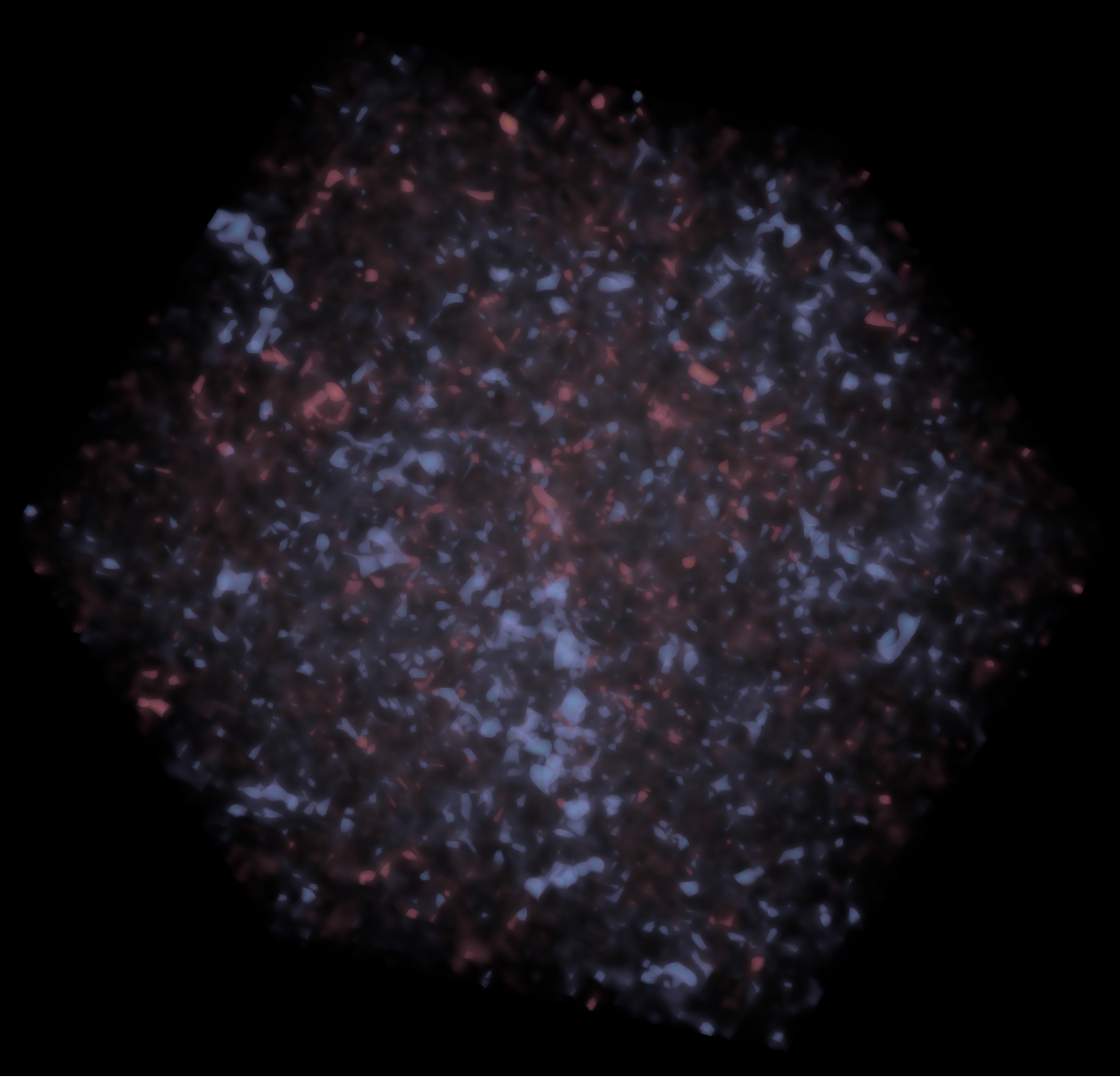}
        \caption{$ \boldsymbol \Phi_3^4$ at $t=1$}
        \label{fig:phi43-c}
    \end{subfigure}
    \caption{The dynamical \eqref{eq:phi43} model at various times $t$, beginning at white noise initial condition.}
    \label{fig:phi43-triptych}
\end{figure}
To simulate the dynamic $\boldsymbol{\Phi}^4_3$ model, we follow the renormalised lattice approximation of \citet{zhu2018lattice}, in which the continuum SPDE is replaced by a finite-dimensional SDE on a periodic three-dimensional lattice $\Lambda_\epsilon \subset \mathbb{T}_L^3$.

Specifically, on a finite time interval $[0,T]$ and the periodic lattice $\Lambda_\epsilon$, we consider
\begin{equation}
    \mathrm{d}\Phi^\epsilon(t,x)
    =
    \Bigl(
        \Delta_\epsilon \Phi^\epsilon(t,x)
        -
        \bigl(\Phi^\epsilon(t,x)\bigr)^3
        +
        \bigl(3C_0^\epsilon - 9C_1^\epsilon\bigr)\Phi^\epsilon(t,x)
    \Bigr)\,\mathrm{d}t
    +
    \mathrm{d}W^\epsilon(t,x),
    \tag{$\boldsymbol{\Phi}_3^4$}
    \label{eq:phi43}
\end{equation}
where $\Delta_\epsilon$ is the nearest-neighbour discrete Laplacian with periodic boundary conditions, and the renormalisation enters through the mass shift $3C_0^\epsilon - 9C_1^\epsilon$. In the Zhu--Zhu formulation, $C_1^\epsilon$ further decomposes into a principal $P_N$ contribution and bounded sideband corrections associated with $\Pi_N$.

Numerically, we do not attempt to discretise the paracontrolled formulation itself. Rather, we simulate the renormalised lattice SDE by a first-order semi-implicit Euler scheme. Writing $\lambda_\epsilon(k)$ for the Fourier symbol of $-\Delta_\epsilon$,
\[
    \lambda_\epsilon(k)
    =
    \frac{4}{\epsilon^2}
    \sum_{j=1}^3
    \sin^2\!\left(\frac{k_j\epsilon}{2}\right),
    \qquad
    \widehat{\Phi}^{\,n+1}(k)
    =
    \frac{\widehat{R}^{\,n}(k)}{1 + \Delta t\,\lambda_\epsilon(k)},
\]
so the stiff linear part is treated implicitly in Fourier space via FFT diagonalisation. The real-space right-hand side is
\[
    R^n
    =
    \Phi^n
    +
    \Delta t\Bigl(
        -(\Phi^n)^3
        +
        m_\epsilon \Phi^n
    \Bigr)
    +
    \Delta W^n,
    \qquad
    m_\epsilon \coloneqq 3C_0^\epsilon - 9C_1^\epsilon,
\]
hence the cubic drift and renormalisation term are applied pointwise in physical space, while $\Delta W^n$ is sampled as i.i.d.\ Gaussian lattice noise with variance $\Delta t\,\epsilon^{-3}$ per site.

For the counterterm, we compute $C_0^\epsilon$ from the lattice spectral sum over nonzero modes, and approximate the principal contribution $C_{11}^\epsilon$ of $C_1^\epsilon$ by FFT-accelerated quadrature in the auxiliary time variable appearing in \citet{zhu2018lattice}. At present, however, we set the bounded sideband contribution $C_{12}^\epsilon$ to zero. Accordingly, the implemented mass shift should be regarded as a partial realisation of the full Zhu--Zhu counterterm rather than the exact renormalisation used in the convergence theorem.

In the final version, we plan to implement the missing counterterm and move to a stochastic exponential Runge–Kutta scheme, to ensure that the simulated data are of adequate numerical quality for the experiments.

The main empirical question is whether methods that better respect the analytic structure of singular SPDEs yield measurable gains on this problem. In particular, $\boldsymbol{\Phi}^4_3$ provides a stringent test of long-time stability, robustness under lattice refinement, and cross-resolution generalisation in a genuinely renormalised stochastic dynamics.

\section{Conclusion}
In this work, we present WCE-FiLM-NO, one of the prize-winning models in the singular SPDE modelling competition. WCE-FiLM-NO follows the spirit of DPDD by firstly learning the solution of the shift equation by inserting sufficient Wick-Hermite features to FNO, whose outputs are further adjusted by affine transformations that are parameterised by simple 2D convolution. WCE-FiLM-NO shows supreme results via different noise spectral truncations and cross-truncation evaluations. We also develop the simulation method for more complicated $\boldsymbol{\Phi}^4_3$, providing extra space and challenge for future learning tasks. 


\bibliographystyle{plainnat}
\bibliography{ref}

\end{document}